\documentclass[conference]{IEEEtran}
\IEEEoverridecommandlockouts

\def\BibTeX{{\rm B\kern-.05em{\sc i\kern-.025em b}\kern-.08em
    T\kern-.1667em\lower.7ex\hbox{E}\kern-.125emX}}
\usepackage[boxed, ruled, linesnumbered]{algorithm2e}
\usepackage{multirow}
\usepackage{amssymb}
\usepackage{amsmath}
\DeclareMathOperator*{\argmin}{arg\,min}

\usepackage{mathtools}
\usepackage{nccmath}
\usepackage{subfiles}
\usepackage{bm}
\usepackage{multicol}

\usepackage{subcaption}
\usepackage{graphicx}
\usepackage{wrapfig}
\usepackage{booktabs}

\usepackage{tikz}
\usetikzlibrary{arrows, automata, positioning}

\title{Metagame Autobalancing for Competitive Multiplayer Games
\thanks{Thank you, Ozan Vardal, Nick Ballou and Sebastian Berns for your generous help in coding Workshop Warfare. This work was funded by the EPSRC Centre for Doctoral Training in Intelligent Games and Game Intelligence (IGGI) EP/L015846/1 and Digital Creativity Labs.}}

\author{
    \IEEEauthorblockN{Daniel Hernandez\IEEEauthorrefmark{1}, Charles Takashi Toyin Gbadamosi\IEEEauthorrefmark{2}, James Goodman\IEEEauthorrefmark{3}\\ and James Alfred Walker\IEEEauthorrefmark{1},~\IEEEmembership{Senior Member, IEEE}}\\

\IEEEauthorblockA{\IEEEauthorrefmark{1}Department of Computer Science, University of York, UK.~\{dh1135, james.walker\}@york.ac.uk}
\IEEEauthorblockA{\IEEEauthorrefmark{2}Department of Computer Science, Queen Mary University of London, UK.~\{c.t.t.gbadamosi, james.goodman\}@qmul.ac.uk}
}

\begin{document}
\newcommand{\vertdots}{\underset{\big{\overset{\cdot}{\cdot}}}{\cdot}} 
\newcommand{\diagdots}{_{^{\big\cdot}\cdot _{\big\cdot}}}

\IEEEpubid{\begin{minipage}{\textwidth}\ \\[12pt]
978-1-7281-4533-4/20/\$31.00 \copyright 2020 Crown
\end{minipage}}

\maketitle

\begin{abstract}
Automated game balancing has often focused on single-agent scenarios. In this
paper we present a tool for balancing multi-player games during game design.
Our approach requires a designer to construct an intuitive graphical
representation of their meta-game target, representing the relative scores that
high-level strategies (or decks, or character types) should experience. This
permits more sophisticated balance targets to be defined beyond a simple
requirement of equal win chances. We then find a parameterization of the game
that meets this target using simulation-based optimization to minimize the
distance to the target graph. We show the capabilities of this tool on examples
inheriting from Rock-Paper-Scissors, and on a more complex asymmetric fighting
game.

\end{abstract}

% What were the limitations of previous approaches.
% Why my approach is a ``fresh'' approach. Doing something different to previous things. Stuff before doesn't look into multiplayer. Talk about current ways of game designer (problems and challenges) VS our approach (emphasize how it uses a graphical representation). Barrier of entry to game designer should be low. Our approach is an accessible, non-technical way (Game designer terminology) We present the backend for that system.
% Focus on the fact that this is a tool for game designers. Pitch and presentation is key.

\section{Introduction}

% https://gaigresearch.github.io/gvgaibook/PDF/chapters/ch07.pdf?raw=true
% Introduction to automatic game tuning

Achieving game balance is a primary concern of the game designer, but balancing
games is a largely manual process of trial and error. This is especially
problematic in asymmetric multiplayer games where perceived fairness has a
drastic impact on the player experience. Changes to individual game elements or
rules can have an impact on the balance between high-level strategies that
depend on these, but this impact is unknown before changes are made and can
only be guessed at by designers through experience and intuition. We term this
balance between emergent high-level strategies the `Meta-game balance'. While
in-house tools can be built for the adjustment and authoring of individual game
elements. There are no tools for balancing and adjusting -game elements.

Game balancing takes a lot of time and resources, with current trends
indicating a systematic increase in the cost of game
development~\cite{shaker2016procedural}. It is reliant on human intuition and expert knowledge to estimate how changes in the game mechanics affect emergent gameplay.
% always boilsdown to expert knowledge.
Human play testing as part of this process is time consuming, requiring
many human testers for long play-sessions, which grow longer with more complex
games. In short, human play testing does not scale. 
%We present a simulation-based approach to balancing multiagent games.

An alternative approach to the discovery of meta-game changes that arise from game changes is through data analytics. Large scale multiplayer titles that have
access to large quantities of player data can use a variety of techniques
% the work by Choong-Soo Lee and Ivan Ramler "Investigating the Impact of Game Features on Champion Usage in League of Legends"
to make judgements about the
state of the meta-game and provide designers with insight into future
adjustments, such as \cite{lee2015investigating}.

There are, however, several problems with this approach.
Analytics can only discover balance issues in content that is live, and by that
point balance issues may have already negatively impacted the player
experience: this is a reactive approach and not a preventive one. Worse, games which do not have access to large volumes of player data - less
popular games - cannot use this technique at all.

Furthermore, the process of data analytics itself is not typically within the
skill-set of game designers. It is common for studios that run multiplayer
games to hire data scientists to fill this need. This, in combination with the
trial and error nature of the balance process, results in increased costs,
becoming as a bottleneck for the development of new content.

The importance of meta-game balance and the aforementioned issues motivate
alternate approaches to game balance. This paper presents one such alternative - an automated simulation-based approach to meta-game balance of multiplayer games.
Our approach allows designers to directly
specify a meta-game balance state and have the game parameters that
would create the desired meta-game be discovered automatically by a group of
agents.

% TODO: find way of putting this in

\section{Preliminary notation}\label{section:preliminary_notation}

Cursive lowercase letters represent scalars ($n$). Bold lowercase, vectors ($\bm{\pi} \in \mathbb{R}^n$). Bold uppercase, matrices ($\bm{A} \in \mathbb{R}^{n \times n}$).

\vspace{-1.05em}
\subsection{Game parameterization}

Every video game presents a (potentially very large) number of values that
characterize the game experience, which we shall refer to as \textbf{game
parameters}. These values can be numerical (such as gravitational strength,
movement speed, health) or categorical (whether friendly fire is activated, to
which team a character belongs). As a designer, choosing a good set of
parameters can be the difference between an excellent game and an unplayable
one. We let $E_{\bm{\theta}}$ denote a game environment, parameterized by an
$n$-sized parameter vector $\bm{\theta} \in \{ \Pi_{ i \leq n} \Theta_i \}$,
where $\{ \Pi_{ i \leq n} \Theta_i \}$ represents the joint parameter space,
and $\Theta_i$ the individual space of possible values for the $i$th parameter
in $\bm{\theta}$.

%\subsection{Normal form games}
%
%A \textbf{normal form game}, also known as strategic form games, is a tuple
%($\bm{S}$, $U$, $n$) where $n$ is the number of players, $\bm{S} = (S_1,
%\ldots, S_n)$ is the joint strategy set, where $S_i$ represents the set
%of (pure) strategies available to player $i$. $U: S \rightarrow \mathbb{R}^n$
%is a payoff table mapping each joint action to a scalar score (i.e utility) for
%each player.
% % TODO: consider introducing extended form games as a low level representation of a game
% % and normal form games as a swhether to talk / mention about extended form games.
% % JNG: Personally I would remove this subsection. It made more sense when we just had RPS
% % in the paper, but why go to this level of detail for RPS, but not for 'Workshop Warfare'?
% % I don't feel this is needed of the later discussion of RPS, and takes up valuable space?
%
%For simplicity, we will concern ourselves with 2-player zero-sum games. These
%are represented by a single matrix $\bm{M}$. The entry $m_{ij} \in
%\bm{M}$ represents the payoff for player 1 (and $-a_{ij}$ for player 2) when
%player 1 chooses strategy $i$ and player 2 chooses strategy $j$.
%
\subsection{Meta-games}
What a \textbf{meta-game} is can mean different things to different players.
For example in deck-building games such as Hearthstone, the `meta' is usually
interpreted to indicate which decks are currently popular or especially strong;
while in EVE Online an important part of the `meta' is player diplomatic
alliances, as well as which ship types are good against which others. See
\cite{carter2012metagames} for a good discussion of this notation.

In this work we define a meta-game as a set of high-level strategies that are
abstracted from the atomic game actions. Reasoning about a game involves
thinking about how each individual action will affect the outcome of the game.
In contrast, a meta-game considers more general terms, such as how an
aggressive strategy will fare against a defensive one. In meta-games,
high level strategies are considered instead of primitive game actions. Take a
card game like Poker. Reasoning about a Poker meta-game can mean reasoning
about how bluff oriented strategies will deal against risk adverse strategies.

The level of abstraction represented in a meta-game is defined by the
meta-game designer, and the same game can allow for a multitude of different
 levels of abstraction. For instance, in the digital card game of Hearthstone,
meta-strategies may correspond to playing different deck types, or whether to
play more offensively or defensively within the same deck. 
A game designer may want to ensure that no one deck type dominates, but be happy that a particular deck can only win if played offensively.

\subsection{Empirical win-rate matrix meta-games}

An interesting meta-game definition that has recently received 
attention in multiagent system analysis~\cite{omidshafiei2019alpha} defines a normal form game over a
population of agents $\bm{\pi}$, such that the action set of each player
corresponds to choosing an agent $\pi_i \in \bm{\pi}$ from the population to
play the game for them. How these agents were created is not relevant to us;
these agents could use hand-crafted heuristics, be trained with
reinforcement learning, evolutionary algorithms or any other method.
%for developing game-playing agents.

Let $\bm{W}_{\bm{\pi}} \in \mathbb{R}^{n \times n}$ denote an \textbf{empirical
win-rate matrix}. The entry $w_{i,j}$ for $i,j \in \{n\}$ represents the win-rate
of many head-to-head matches of policy $\pi_i$ when playing against policy
$\pi_j$ for a given game. An empirical win-rate matrix $\bm{W_{\pi}}$ for a
given population $\bm{\pi}$ can be considered as a payoff matrix for a 2-player
zero-sum game. An empirical win-rate matrix can be defined over two
(or more) populations $\bm{W}_{\bm{\pi_1}, \bm{\pi_2}}$, such that each player
chooses agents from a different population. 
We can investigate the strengths and weaknesses of each these agents in this kind of meta-game using
game-theoretical analysis.

An \textbf{evaluation matrix}~\cite{Balduzzi2019} is a generalization of an
empirical win-rate matrix. Instead of representing the win/loss ratio
between strategies, it captures the payoff or score obtained by both the
winning and losing strategy. That is, instead of containing win-rates for a
given set of agents, an entry in an evaluation matrix $a_{ij} \in \bm{A}$ can
represent the score obtained by the players.

\subsection{Empirical Response Graphs}

% Discrete-time dynamics \cite{omidshafiei2019alpha}
% Empirical game deviation graphs \cite{wellman2006}

A \textbf{directed weighted graph} of $v \in \mathbb{N}^+$ nodes can be denoted
by an adjacency matrix $\bm{G} \in \mathbb{R}^{v \times v}$. Each row $i$ in
$\bm{G}$ signifies the weight of all of the directed edges stemming from node
$i$. Thus, $g_{i,j} \in \mathbb{R}^+$ corresponds to the weight of the edge
connecting node $i$ to node $j$, where $ 1 \leq i, j \leq v$.

Given an evaluation matrix $\bm{A_{\pi}}$ computed from a set of strategies (or
agents) $\bm{\pi}$, let its \textbf{response graph}~\cite{wellman2006}
represent the dynamics~\cite{omidshafiei2019alpha} between agents in
$\bm{\pi}$. That is, a representation of which strategies (or agents) perform
favourably against which other strategies in $\bm{\pi}$. In a competitive scenario,
a response graph shows which strategies win against which others. As a
graph, each strategy $i$ is represented by a node. An edge connecting node
$i$ to node $j$ indicates that $i$ dominates $j$. The weight of the edge
is a quantitative metric of how favourably strategy $i$ performs against $j$.
Figure~\ref{fig:RPS_response_graph} shows a response graph for the game of
Rock-Paper-Scissors (RPS).

A response graph can be readily computed from an evaluation matrix. Each row
$i$ in an evaluation matrix $\bm{A}$ denotes which strategies $i$ both
wins and loses against, the former being indicated by positive entries and the
latter by negative ones. Therefore, generating a response graph $\bm{G}$ from
an evaluation matrix $\bm{A}$ is as simple as setting all negative entries of
$\bm{A}$ to $0$ such that, for instance, $\bm{A} =
\begin{psmallmatrix*}[r] 1 & -2\\ 2 & -1\end{psmallmatrix*}$, becomes $\bm{G} =
\begin{psmallmatrix*}[r] 1 & 0\\ 2 & 0\end{psmallmatrix*}$.

% it's essence, a response graph shows which strategies beat which other strategies. Each row $i$ in a response graph $\bm{G}$ denotes which strategies does $i$ beat. In contrast, 

\subsection{Graph distance}

There is a rich literature on measuring distance between
graphs~\cite{gao2010}. We concern ourselves here with a basic case. We are
interested in measuring the distance between two graphs which share the same number
of nodes, $\bm{G_1}, \bm{G_2} \in \mathbb{R}^{v \times v}$, and differ
\textit{only} in the weight of the edges connecting nodes. Because graphs
can be represented as matrices, we look at differences between matrices.
We denote the distance between two graphs $\bm{G^1}$ and $\bm{G^2}$ by
$d(\bm{G^1}, \bm{G^2}) \in \mathbb{R}$. Equation (1) represents the average absolute edge
difference and (2) represents the mean squared difference (MSE).

\begin{multicols}{2}
  \centering
  \begin{equation}
      \frac{\sum_{i,j} |g^1_{ij} - g^2_{ij}|}{n}
  \end{equation}
  \begin{equation}
      \frac{\sum_{i,j} (g^1_{ij} - g^2_{ij})^2}{n}
      \label{eq:MSE}
  \end{equation}
\end{multicols}

Preliminary results showed no empirical difference between distance metrics (1)
and (2). Thus, we report only the results where MSE (Equation~\ref{eq:MSE}) was used.

\section{Autobalancing}\label{section:balancing}

In this section we present our autobalancing algorithm in its most general form.

\subsection{Optimization setup}

% Each node in $\bm{G_t}$ corresponds to a unit in the game.

Let $E_{\bm{\theta}}$ be a game environment parameterized by vector
$\bm{\theta} \in \mathbb{R}^n$, whose possible values are bound by vectors
$\bm{\theta}^{min}$ and $\bm{\theta}^{max}$. Let $\bm{G_t}$ denote the target
metagame response graph presented by a game designer for game
$E_{\bm{\theta}}$. Let $\bm{G_{\theta}}$ represent the empirical metagame
response graph produced from a set of gameplaying agents $\bm{\pi}$ for game
$E_{\bm{\theta}}$, where each agent corresponds to a node in the graph
$\bm{G_{\theta}}$. Finally, let $\mathcal{L}(\cdot, \cdot)$ represent a cost or
distance function between two graphs.

The mathematical formulation for finding a parameter vector $\bm{\theta}$ which
yields a metagame for a game environment $E_{\bm{\theta}}$ respecting designer
choice $\bm{G_t}$ is a constrained non-linear optimization problem:

\begin{align}
    \argmin_{\bm{\theta}} &\quad \mathcal{L}(\bm{G_{\theta}}, \bm{{G_t}}) \\
                 s.t &\quad \theta^{min}_i \le \theta_i \le \theta^{max}_i \; \forall i \in \{|\bm{\theta}|\}
    \label{equation:optimization-autobalancing}
\end{align}

\begin{algorithm}
    \KwIn{\textit{Target designer meta-game response graph}: $\bm{G_t}$}
    \KwIn{\textit{Ranges for each parameter}: $\bm{\theta}^{min}, \bm{\theta}^{max}$}
    \KwIn{\textit{Convergence threshold}: $\epsilon$}
    Initialize game parameterization $\bm{\theta_0}$\;
    Initialize best estimate $\bm{\theta}_{best}, \mathcal{L}_{best} = \bm{\theta_0}, \infty$\;
    Initialize observed datapoints $D = [\ ]$\;
    \Repeat{$\mathcal{L}(\bm{G_{\theta_t}}, \bm{G_t}) < \epsilon$} {
        Train agents $\bm{\pi}$ inside $E_{\bm{\theta_t}}$, for each node in $\bm{G_t}$\;
        Construct evaluation matrix $\bm{A_{\theta_t}}$ from $\bm{\pi}$\;
        Generate response graph $\bm{G_{\theta_t}}$\;
        Compute graph distance $d_t = \mathcal{L}(\bm{G_{\theta_t}}, \bm{G_t})$\;
        Add new datapoint $D = D \cup (\bm{\theta_t}, d_t)$\;
        \If{$d_t < \mathcal{L}_{best}$}{
            Update best estimate $\bm{\theta}_{best}, \mathcal{L}_{best} = \bm{\theta_t}, d_t$\;
        }

        $\bm{\theta_{t+1}} = update(\bm{\theta_{t}}, D)$\;
    }
    return $\bm{\theta}_{best}$\;

    \caption{Automated balancing algorithm.}\label{algorithm:automated_balancing}
\end{algorithm}

There are four notes to be made about our algorithm:

\begin{enumerate}
    \item \textbf{It can be parallelized}: multiple parameter vectors can be evaluated simultaneously.
    \item \textbf{It allows for initial designer choice}: such that designers can designate an initial parameter vector and a prior over the search space, which can lead to speedup in the convergence of the algorithm.
    \item \textbf{An arbitrary subset of the game parameters can be fixed}: $\bm{\theta}$ can represent a subset of the entire game parameters. This is important if there are certain core aspects of a game that the designer does not want to be altered throughout the automated game balancing.
    \item \textbf{Deterministic results are not guaranteed}. There are three potential sources of stochasticity, the game dynamics $E_{\bm{\theta}}$, the agent policies $\bm{\pi}$ and the optimizer's parameter choices (line 13 of Algorithm~\ref{algorithm:automated_balancing}).
\end{enumerate}

There are two potential bottlenecks in Algorithm $1$ in terms of the
computational requirements of (1) the construction of the evaluation matrix and
(2) the update of the parameter vector. The main computational burden in (1)
comes from the fact that computing each entry in an evaluation matrix $a_{ij}
\in \bm{A}$ require running many game episodes played by agents $i$ and $j$,
with the cost of computing $\bm{A}$ growing exponentially with respect to the
number of agents.

\subsection{Choosing an optimizer}

We want to emphasize that our algorithm can use any black-box optimization
method. To compute updates to our parameter vector $\bm{\theta}$ we use
Bayesian optimization. Specifically, we use the algorithm Tree-structured
Parzen Estimator~\cite{bergstra2011}, as implemented in the Python framework
Optuna~\cite{optuna}, but this could be replaced with any other optimization
method. Most commonly in the literature of automated game balancing,
evolutionary algorithms have been
used~\cite{morosan2017evolving}.

\subsection{Choosing a metagame abstraction}

For most games, there are many possible abstractions (and levels of
abstraction) available when deciding what the metagame captured by the target
response graph represents.

Choosing the abstraction may not be obvious, but
we argue that reasoning about metagames is a necessary task in balancing any
multi-agent game. On a positive note, the fact that metagames can be represented
at many levels of abstraction grants our method the versatility to generalize
to various stages of balancing. That is to say, our method can be used at
different points of game development to balance different aspects of the game.

Generally each node on the response graph represents a specific strategy, unit
or game-style. A possible target response graph could symbolize
the interactions between players or agents trained to represent different
in-game ``personas''~\cite{bartle2008player}, where a ``persona'' representing
a different reward scheme for an agent. In an RPG each node of the response
graph might represent a character class; Paladin, Wizard, Sniper etc., as we
seek to balance these against each other. At a lower level, each node might
represent an individual weapon.

During auto-balancing we train an AI to play each of the strategies/units that
the nodes represent as well as possible, where this will often mean `winning',
but could use some other balance target such as `gold gained', or `length of
fight'.

\subsection{Generating game-playing agents}

As specified in Section~\ref{section:preliminary_notation}, in order to compute
an evaluation matrix $\bm{A}$ for a given game $E_{\bm{\theta}}$ we require a
set of gameplaying agents $\bm{\pi}$. These could be hand-crafted heuristic
agents, or agents trained via reinforcement learning or evolutionary
algorithms~\cite{Hernandez2019, yannakakis2018artificial}. The algorithmic
choice for how to train these agents is orthogonal to the usage of our method.
However, we acknowledge that the creation of these agents can be a significant
engineering and technical effort.

\section{Motivating examples}\label{section:motivating_examples}

% TODO: mention that in these examples we assume that we have a vector of agents
% $\bm{\pi} = [\pi_{rock}, \pi_{paper}, \pi_{scissors}]$.

In this section we present basic examples of our automated balancing algorithm.
For simplicity, we assume that all parameters in the
following examples are bound between $[-1, +1]$. As a graph distance metric we
use $\mathcal{L}(\cdot, \cdot) = MSE(\cdot, \cdot)$ from Equation~\ref{eq:MSE}.

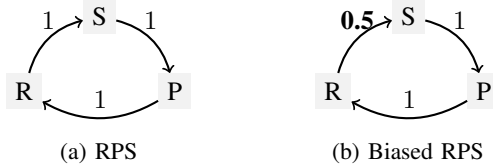
\begin{figure}[htbp]
  \begin{subfigure}[b]{0.45\linewidth}
    \centering
    \begin{tikzpicture}[->]
        \node (R)[fill=gray!10,align=left] at (0,0)  {R};
        \node (S)[fill=gray!10,align=left] at (1,1)  {S};
        \node (P)[fill=gray!10,align=left] at (2,0)  {P};

        \path [bend left, thick] (R) edge node [anchor=center, above] {$1$} (S);
        \path [bend left, thick] (S) edge node [anchor=center, above] {$1$} (P);
        \path [bend left, thick] (P) edge node [anchor=center, above] {$1$} (R);

    \end{tikzpicture}
    \caption{RPS}
    \label{fig:RPS_response_graph}
  \end{subfigure}
  \begin{subfigure}[b]{0.45\linewidth}
    \centering
    \begin{tikzpicture}[->]
        \node (R)[fill=gray!10,align=left] at (0,0)  {R};
        \node (S)[fill=gray!10,align=left] at (1,1)  {S};
        \node (P)[fill=gray!10,align=left] at (2,0)  {P};

        \path [bend left, thick] (R) edge node [anchor=center, above] {$\textbf{0.5}$} (S);
        \path [bend left, thick] (S) edge node [anchor=center, above] {$1$} (P);
        \path [bend left, thick] (P) edge node [anchor=center, above] {$1$} (R);

    \end{tikzpicture}
    \caption{Biased RPS}
    \label{fig:biased_RPS_response_graph}
  \end{subfigure}
  \caption{Target graphs for the 2 motivational examples}
  \label{fig:RPS_response_graphs}
\end{figure}

\subsubsection{Rock Paper Scissors}

% How do we mention that all interactions are equally strong?
%Initially we want to create a fully cyclic game (all strategies are equally good~\cite{Balduzzi2019}), and thus all of the edges of Figure~\ref{fig:RPS_response_graph} are labeled with a $1$.
Imagine we want to create the game of Rock Paper Scissors
\footnote{https://en.wikipedia.org/wiki/Rock\%E2\%80\%93paper\%E2\%80\%93scissors}.
As a designer choice, we want paper to beat rock, rock to beat scissors and
scissors to beat paper, with mirror actions negating each other. Such strategic
balancing is captured in Figure~\ref{fig:RPS_response_graph}. These
interactions can be represented as a 2-player, symmetric, zero-sum normal form
game $\bm{E}^{RPS}_{\bm{\theta}}$, parameterized by $\bm{\theta} =
[\theta_{rp}, \theta_{rs}, \theta_{ps}]$. Where $\theta_{rp}$ denotes the
payoff for player 1 when playing Rock against Paper, $\theta_{rs}$ when playing
Rock against Scissors and $\theta_{ps}$ when playing Paper against Scissors.
The normal form parameterized version of RPS is captured in
Equation~\ref{eq:RPS_parameterized}. We ask the question: Which parameter
vector $\bm{\theta}$ would yield a game $\bm{E}^{RPS}_{\bm{\theta}}$ balanced
as in Figure~\ref{fig:RPS_response_graph}?.

\begin{equation}
  \bm{E}^{RPS}_{\bm{\theta}} = 
  \begin{bmatrix}
      \cdot & R & P & S \\
      R & 0 & \theta_{rp} & \theta_{rs}\\
      P & - \theta_{rp} & 0 & \theta_{ps}\\
      S & - \theta_{rs} & - \theta_{ps} & 0
  \end{bmatrix}
  \label{eq:RPS_parameterized}
\end{equation}

\begin{wrapfigure}{R}{0.13\textwidth}
\centering
    \begin{tikzpicture}[->]
        \node (R)[fill=gray!10,align=left] at (0,0)  {R};
        \node (S)[fill=gray!10,align=left] at (1,1)  {S};
        \node (P)[fill=gray!10,align=left] at (2,0)  {P};

        \path [bend left, thick] (R) edge node [anchor=center, above] {$1$} (S);
        \path [bend left, thick] (P) edge node [anchor=center, above] {$1$} (R);

    \end{tikzpicture}
    \caption{} % Intermediate response graph for $\bm{E_{\theta_0}}$}
    \label{fig:RPS_intermediate_response_graph}
\end{wrapfigure}

We begin by assuming the target balance response graph $\bm{G_t}$ from
Figure~\ref{fig:RPS_response_graph} is given by a game designer. Lacking any
informed priors, we start by sampling a random valid parameter vector, say,
$\bm{\theta_0} = [-1, 1, 0]$. We then generate an evaluation matrix by pitting
Rock, Paper and Scissors against each other, yielding $\bm{A_{\theta_0}} =
\begin{psmallmatrix*}[r]0 & -1 & 1\\ 1 & 0 & 0\\ -1 & 0 &
0\end{psmallmatrix*}$, whose response graph $ \bm{G_{\theta_0}} =
\begin{psmallmatrix*}[r]0 & 0 & 1\\ 1 & 0 & 0\\ 0 & 0 & 0\end{psmallmatrix*}$
is depicted in Figure~\ref{fig:RPS_intermediate_response_graph}. We proceed by
computing the distance between $\bm{G_{\theta_0}}$ and $\bm{G_{t}}$,
$d_{\bm{\theta_0}} = \mathcal{L}(\bm{G_{\theta_t}}, \bm{G_t}) = 0.25$. Using
this new datapoint ($\bm{\theta_0}$, $d_{\bm{\theta_0}}$) we update our black
box optimization model, which in our case is Bayesian optimization, and sample
a new $\bm{\theta_1}$. This process is looped until convergence or an arbitrary
computational budget is spent.
% JNG: In Figure 3 I do not understand the need to plot three lines (for initial, intermediate and final values). Why do we not just plot one line per experiment showing the decrease in graph distance as iterations increase?

% TODO: talk about how we would get these values if we didn't know the normal game

\subsubsection{Biased Rock Paper Scissors}

Consider another version of Rock Paper Scissors where we want to
\textit{weaken} the strength of playing Rock, as denoted in
Figure~\ref{fig:biased_RPS_response_graph}. For our algorithm, this amounts to
discovering a lower payoff $\theta_{rp}$ obtained by playing Rock against
Scissors.

\begin{figure}[htbp]
  \begin{subfigure}[b]{0.495\linewidth}
    \centering
    \includegraphics[width=\textwidth,keepaspectratio]{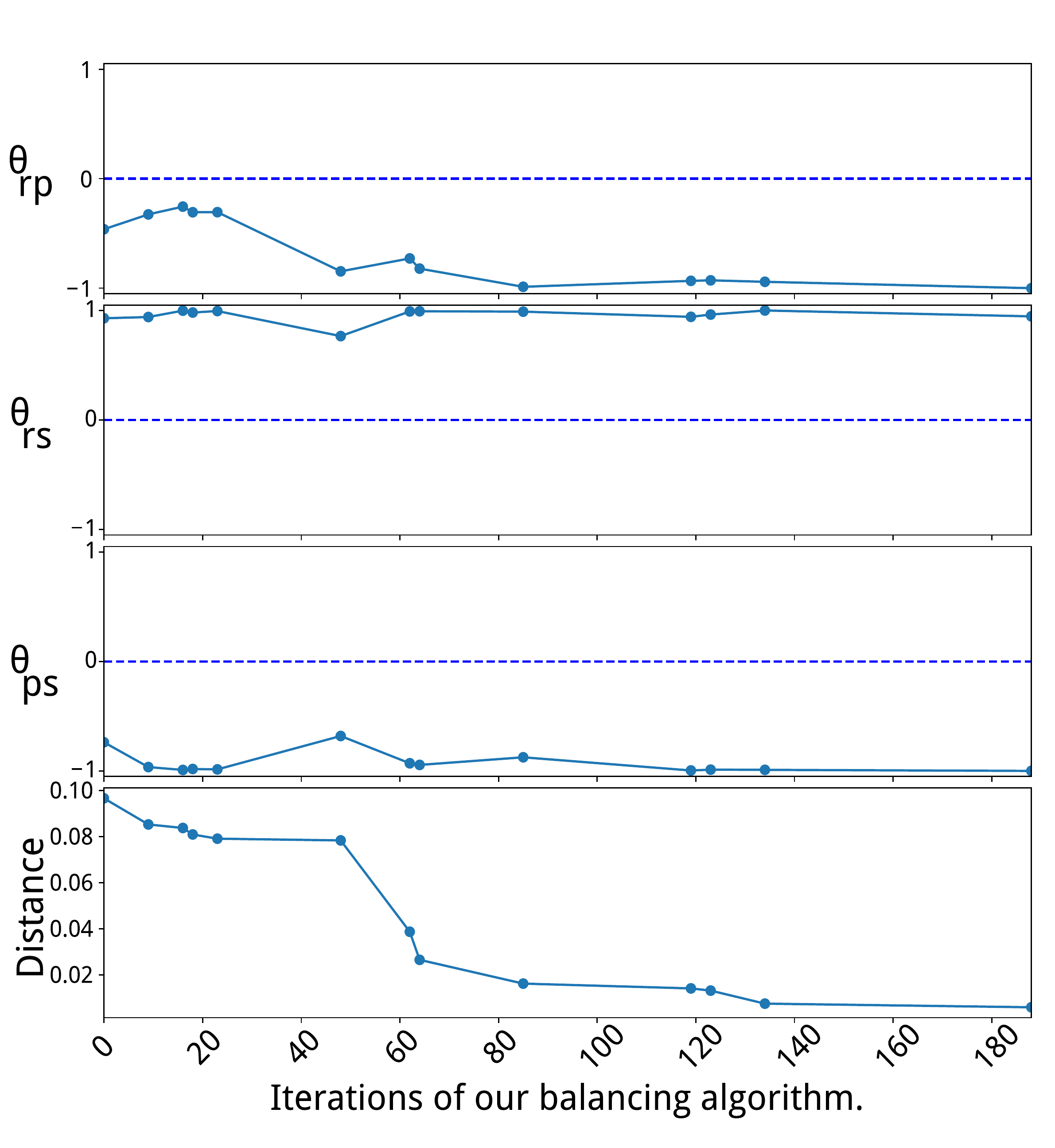}
    \caption{RPS}
    \label{fig:RPS_balancing_progression}
  \end{subfigure}
  \begin{subfigure}[b]{0.495\linewidth}
    \centering
    \includegraphics[width=\textwidth,keepaspectratio]{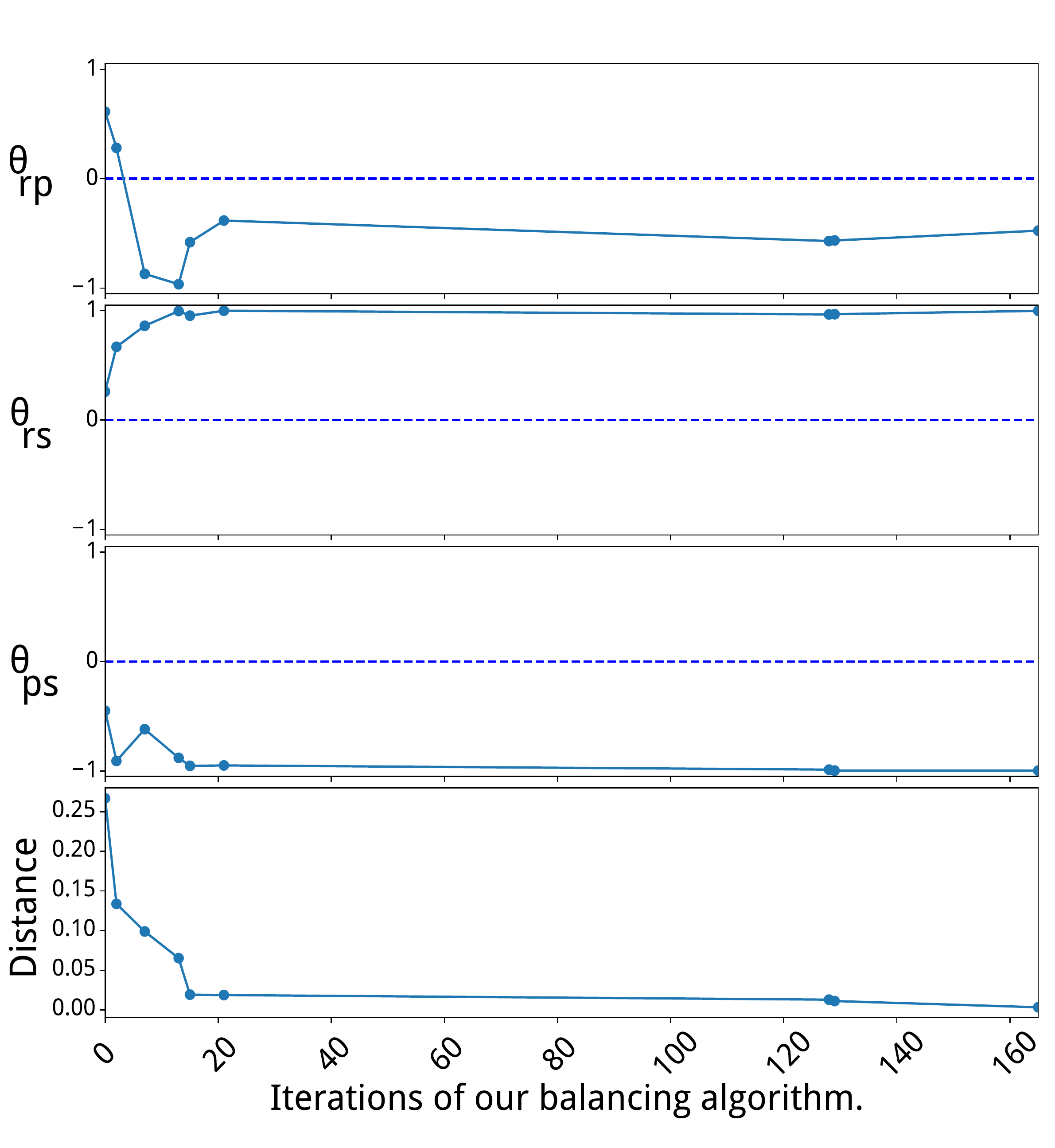}
    \caption{Biased RPS}
    \label{fig:biased_RPS_balancing_progression}
  \end{subfigure}
  %\begin{subfigure}[b]{0.3\linewidth}
  %  \centering
  %  \includegraphics[width=\textwidth,keepaspectratio]{images/extended_rps_balancing_progression.png}
  %  \caption{Extended RPS}
  %  \label{fig:extended_RPS_balancing_progression}
  %\end{subfigure}
  \caption{Progression of balance parameters and distance to target graph. Only parameters which improved with respect to the previous best estimate are plotted. The target parameter values for (a) and (b) respectively are: $[-1, 1, -1]$, $[0.5, 1, -1]$.}
  \label{fig:motivational_example_balancing_progression}
\end{figure}

Figure~\ref{fig:motivational_example_balancing_progression} shows the progression of parameter
values $\bm{\theta}$ computed for problems 1) RPS and 2) Biased RPS, described
above. With $1\%$ tolerance, our method converges to the correct parameter
values within $180$, and $160$ respectively.

\section{Usage on a real game}

The parameters optimized in the previous section directly influenced the payoff
obtained by the agents playing the game. This is not a realistic scenario. The
game parameters that designers can directly change impact game mechanics, which
only indirectly affect the outcome of a game. Therefore, for the remainder of
this section, we don our game designer hat, to showcase a usage of our
algorithm in a realistic challenge.

% things change in the deployment of real world game design.
% In this section we present an experiment which more closely
% a real use case by tuning the parameters of a more complex game. 

\subsection{Workshop Warfare: a more realistic game}

\begin{figure}[htbp]
    \centering
    \begin{subfigure}{.3\linewidth}
        \centering
        \includegraphics[width=0.4\linewidth, keepaspectratio]{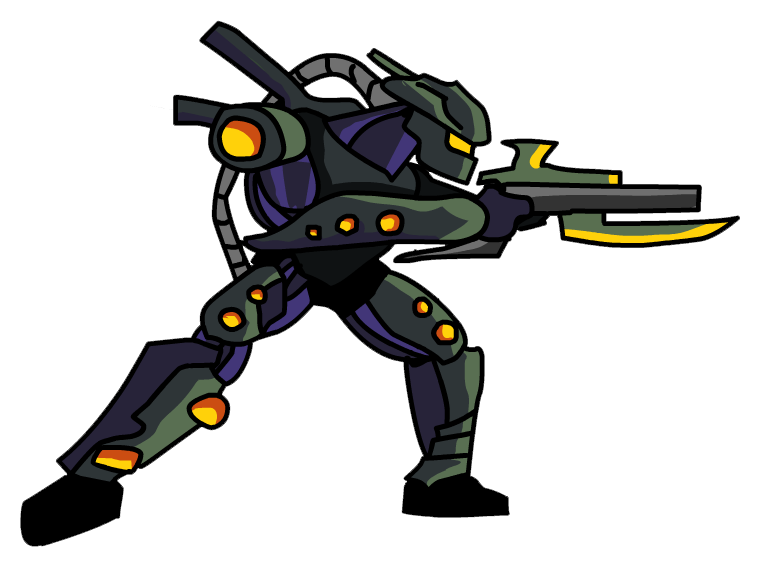}
        \caption{Nail Bot}
        \label{fig:sub1}
    \end{subfigure}%
    \begin{subfigure}{.3\linewidth}
        \centering
        \includegraphics[width=0.4\linewidth, keepaspectratio]{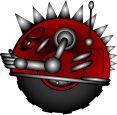}
        \caption{Saw Bot}
        \label{fig:sub1}
    \end{subfigure}%
    \begin{subfigure}{.3\linewidth}
        \centering
        \includegraphics[width=0.4\linewidth, keepaspectratio]{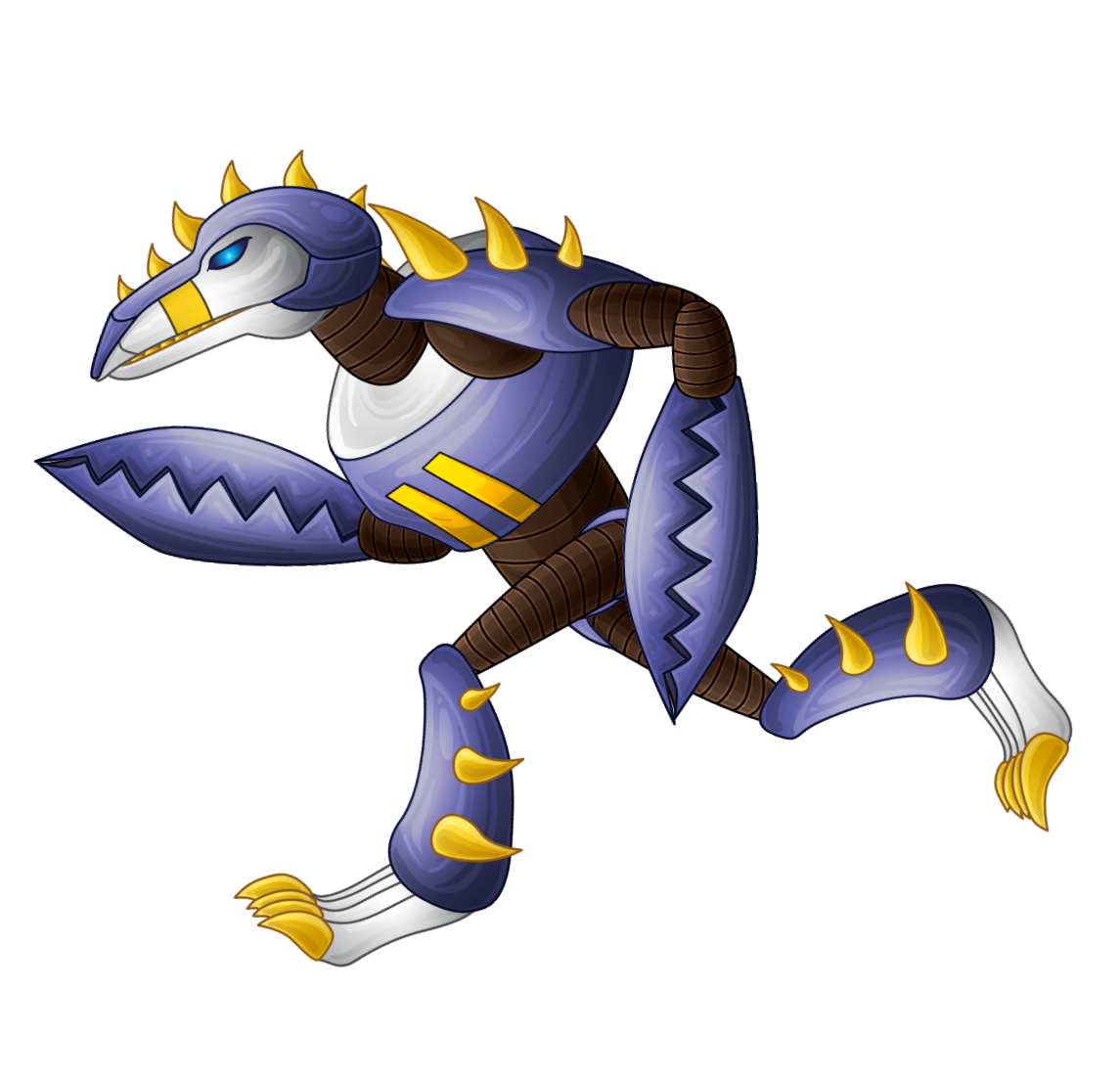}
        \caption{Torch Bot}
        \label{fig:sub1}
    \end{subfigure}%
    \caption{Eligible characters in Workshop Warfare.}
    \label{fig:bots}
\end{figure}

Workshop Warfare\footnote{The game is open source, and follows an OpenAI Gym
interface~\cite{openaigym}: https://github.com/Danielhp95/GGJ-2020-cool-game}
is a 2-player, zero-sum, symmetric, turn based, simultaneous action game. The
theme of the game is a 1v1 battle between robots on a 5x5 grid with the
objective of depleting the opponent's health\footnote{Akin to TV shows like
Battle Bots https://www.wikiwand.com/en/BattleBots}. Each player chooses 1 out
of 3 available robots (Figure~\ref{fig:bots}) to fight the opponent's robot of
choice, with each robot featuring a different style of play. All robots feature
the same action space: standing still (S), moving up (U), down (D), left (L),
right (R) and a special action (A). The special action (A) varies per robot and
will be explained later.

Workshop Warfare works on a ``tick'' basis. Each bot has an associated number
of ticks shared across all action, representing how many in-game ticks must
elapse between actions. This property can be thought as a time cost or robot
``speed''. A bot is said to be ``sleeping'' during the period that it cannot
take actions. Standing still (S) has no cost, meaning that it allows the bot
which took that action to take another action on the next tick. This allows for
a degree of strategic depth.

To clarify the tick based system, imagine an scenario with two bots, with 2 and 4
ticks respectively. They both select a (U) action, moving upwards
by 1 square in the grid. The next tick will elapse without anything happening,
as both bots are ``sleeping''. On the next tick, bot 1 will be able to act
again, followed by another tick with both bots sleeping. On the next tick both
bots will be able to act again. 

There are no time restrictions placed upon the players at the time of selecting
an action. This makes it amenable for forward planning methods that
use a given computational budget to decide on what action to take.
Thus, when autobalancing, this budget can be
scaled without affecting the flow of the game, this is further explained in
Section~\ref{section:mcts}.

We now describe all three bots, whose in-game sprites are shown in
Figure~\ref{fig:bots}. \textbf{Torch bot} is equipped with a damaging blow
torch, and can shoot a continuous beam of fire of limited range in all four
directions for a short amount of time. \textbf{Nail bot} has a nail gun, and
can shoot nails in all four directions at once. When fired, each nail travels
in a fixed direction, at a speed of one grid cell per tick, independently of
the bot's later movement and deals damage if they hit the opponent. \textbf{Saw
bot}'s spikes deal damage by being adjacent to the opponent. Its ability is to
temporarily increase its damage.

\subsection{Game parameterization}

All bots share some common parameters, although their individual values
can\label{section:mcts} differ from bot to bot. Other parameters are bot
specific and relate to a bot's special action (A).

\begin{enumerate}
    \item[] \textbf{Common parameters}
        \begin{itemize}
            \item \textit{Health}: Damage a bot can sustain before being destroyed.
            \item \textit{Cooldown}: After the special action (A) is activated, number of ticks that
                need to elapse before that action can be used again.
            \item \textit{Damage}: Damage dealt by flames (Torch bot),
                nails (Nail bot) or spikes (Saw bot).
            \item \textit{Ticks between moves}: Number of ticks that need to elapse before
                another action can be taken.
        \end{itemize}
    \item[] \textbf{Bot-specific parameters}
        \begin{itemize}
            \item \textit{Torch range}: Length of the blow torch flame, in number of grid squares. (Torch bot)
            \item \textit{Torch duration}: number of ticks the torch flame is active (Torch bot).
            \item \textit{Damage buff}: Temporary change in damage dealt (Saw bot).
            \item \textit{Duration}: Duration of buff to damage (Saw bot).
        \end{itemize}
\end{enumerate}

The parameters that were optimized in Section~\ref{section:motivating_examples}
were real valued ($\mathbb{R}$), whereas all the parameters in this section are
natural numbers ($\mathbb{N}$). The number of parameter combinations inside the
parameter space would be prohibitively time consuming for any human designers
to manually explore. We now apply our autobalancing method to automate this
process.

\begin{figure}[t]
    \centering
    \begin{subfigure}{0.5\linewidth}
        \centering
        \includegraphics[width=\linewidth, keepaspectratio]{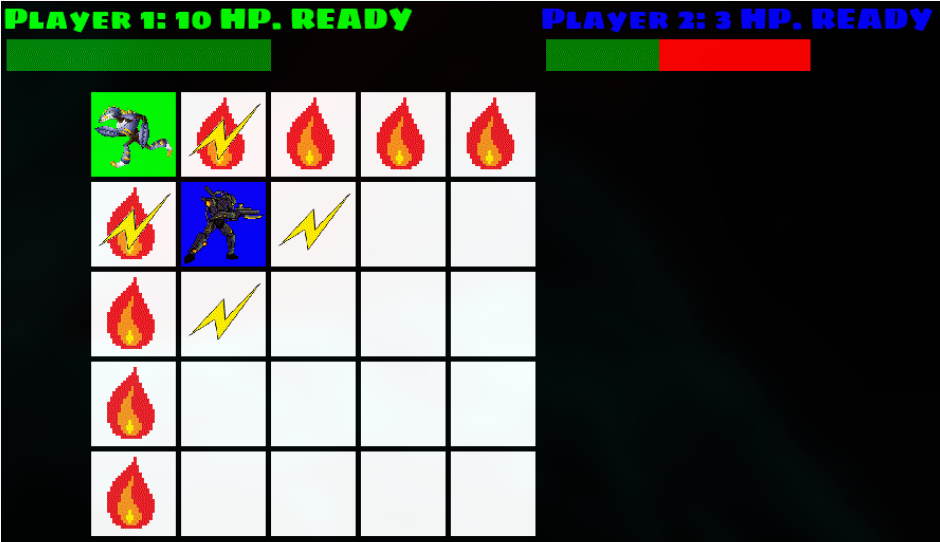}
        \label{fig:sub1}
        \caption{Torch bot VS Saw bot.}
    \end{subfigure}%
    \begin{subfigure}{0.5\linewidth}
        \centering
        \includegraphics[width=\linewidth, keepaspectratio]{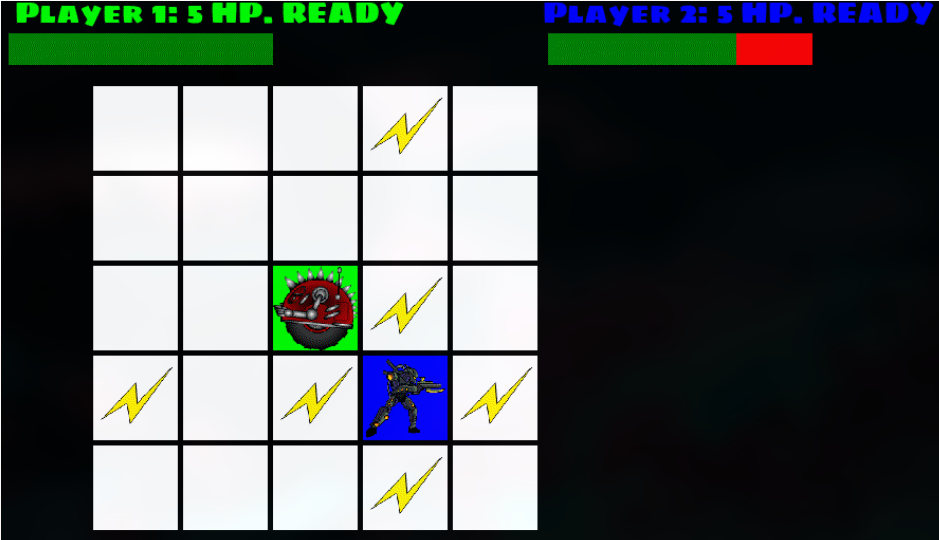}
        \label{fig:sub1}
        \caption{Torch bot vs Nail bot.}
    \end{subfigure}%
    \caption{Screenshots of the game.}
\end{figure}

\section{Experiments on real game}\label{section:experiments_real_game}

We first choose our level of abstraction, and what elements we want to balance
as game designers. For these experiments, we choose to balance the win-rates
between all bot matchups. We want these win-rates to represent the win-rate
between rational competitive players, that is, players who play to win
understanding that their opponent also aims for the same goal. As a proxy of
rational players we use AI agents controlled by Monte Carlo Tree Search (MCTS),
as detailed below.

This level of abstraction differs from the motivating examples from
Section~\ref{section:motivating_examples}. In
Section~\ref{section:motivating_examples} we directly modified the
deterministic payoff obtained by any matchup (i.e, Rock vs Paper, etc). In these
examples we instead aim to balance the win-rate obtained when each bot type is
matched up against each of the other bots. This win-rate is emergent from the
precise parameter settings that we can control, listed in
Table~\ref{table:real_game_params}. This is a much more realistic design
scenario.

\subsection{Target meta-game balance}

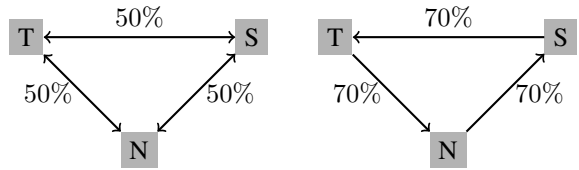
\begin{figure}[h!]
  \begin{subfigure}[b]{0.45\linewidth}
    \centering
    \begin{tikzpicture}[<->]
        \node (T)[fill=gray!60,align=left] at (0,1.5)  {T};
        \node (S)[fill=gray!60,align=left] at (3,1.5)  {S};
        \node (N)[fill=gray!60,align=left] at (1.5,0)  {N};

        \path [thick] (T) edge node [anchor=center, above] {$50\%$} (S);
        \path [thick] (S) edge node [anchor=center, right] {$50\%$} (N);
        \path [thick] (N) edge node [anchor=center, left] {$50\%$} (T);

    \end{tikzpicture}
    \caption{Fair balance}
    \label{fig:cool_game_fair}
  \end{subfigure}
  \begin{subfigure}[b]{0.45\linewidth}
    \centering
    \begin{tikzpicture}[->]
        \node (T)[fill=gray!60,align=left] at (0,1.5)  {T};
        \node (S)[fill=gray!60,align=left] at (3,1.5)  {S};
        \node (N)[fill=gray!60,align=left] at (1.5,0)  {N};

        \path [thick] (S) edge node [anchor=center, above] {$70\%$} (T);
        \path [thick] (T) edge node [anchor=center, left] {$70\%$} (N);
        \path [thick] (N) edge node [anchor=center, right] {$70\%$} (S);

    \end{tikzpicture}
    \caption{Cyclic behaviour}
    \label{fig:cool_game_cyclic}
  \end{subfigure}
  \caption{Target graphs for the 2 experiments on Workshop Warfare. Note the bi-directionality of graph~\ref{fig:cool_game_fair}}
  \label{fig:cool_game_response_graphs}
\end{figure}

We run two experiments, each corresponding to a different design goal.
We will attempt to find the parameter vector $\bm{\theta}$ which yields a
meta-game balance, in terms of bot win-rates, as described by the response
graphs in Figure~\ref{fig:cool_game_response_graphs}. Each element $\theta \in
\bm{\theta}$ corresponds to a game parameter in
Table~\ref{table:real_game_params}.

The two design goals we target are \textit{fair} balancing and \textit{cyclic}
balancing. Fair balancing dictates that all bots should stand an equal chance
of winning against all other bots. All bot win-rates should be $50\%$. Cyclic
balancing dictates that some bots should stand a higher chance at winning
against certain bots than against others. Torch bot should have a $70\%$
win-rate against Nail bot, with the same applying to Nail bot against Saw bot
and Saw bot against Torch bot. This is a relaxed form of Rock-Paper-Scissors,
and will benefit a player able to guess which bot their opponent will choose,
much as in deck selection in a deck-building game.

For these two experiments, the resulting game parameter vectors are shown in
Table~\ref{table:real_game_params}. Given our own computational budget, we let
the \textit{fair} balancing and \textit{cyclic} balancing experiments run for
260 iterations. We ran both experiments on consumer-end
hardware, parallelizing at all times 6 different iterations or trials.

\subsection{Computing an evaluation matrix}

As with Rock Paper Scissors in
Section~\ref{section:motivating_examples}, Workshop Warfare is a 2-player
symmetric zero-sum game. This means that we can exploit the fact that the
win-rate of two bots $a$ and $b$ is $w_{ab} = 1 - w_{ba}$. Let $\bm{\theta}$
denote an arbitrary parameter vector for Workshop Warfare. Let $w^{\bm{\theta}}_{ST}$
denote the win-rate of Saw bot vs Torch bot, Saw bot vs Nail bot
$w^{\bm{\theta}}_{SN}$ and Torch bot vs Nail bot $w^{\bm{\theta}}_{TN}$. Our
algorithm will attempt to find the right set of game parameters $\bm{\theta}$
that yields either a cyclic or fair balancing in terms of these win-rates.

To compute these win-rates, we simulate many head to head matches where each
bot is controlled by an agent using MCTS to guide its actions. Each matchup's
win-rates are computed from the result of 50 game simulations. A higher number
of game simulations would result in a more accurate prediction of the true
win-rate between two bots, at the cost of more computational time.

\subsection{Monte Carlo Tree Search}\label{section:mcts}

A thorough description of MCTS is beyond the scope of this paper,
see~\cite{browne2012survey} for a comprehensive review. Here we use MCTS to
create gameplaying agents\footnote{We have open-sourced our MCTS
implementation: https://www.github.com/Danielhp95/Regym} to auto-balance the
game to meet our design goals.

MCTS relies on a forward-model of the game to run ``internal'' game simulations
alongside the game being played. As such, it would not be possible to use MCTS
on games for which a forward game model is not available, or for which the
model is prohibitively slow. The MCTS agents we use could be replaced with any
other method of creating gameplaying agents suitable for the game of concern,
and are not an integral part of our method.

All MCTS agents use a computational budget of 625 iterations. A higher
computational budget is directly related to a higher skill
level~\cite{Lanzi_2019}. Following this idea, our method could be used to
balance a game at different levels of play by changing the computational
budget.

We use a reward scheme that incentivizes bots to interact with one
another by (1) giving negative score to actions that would increase distance
between bots (2) giving positive / negative score to damaging the opponent /
being damaged and (3) giving a score for winning the game. The magnitude of
rewards (1), (2), and (3) varied between 0-10, 10-99, and 1000 respectively so
as to represent a hierarchy of goals for the agent to follow.

\section{Results}

Using Algorithm~\ref{algorithm:automated_balancing} defined in
Section~\ref{section:balancing}, we found the following parameter vectors
$\bm{\theta}_{fair}$ and $\bm{\theta}_{cyclic}$, corresponding to the meta-game
balancing defined in Figure~\ref{fig:cool_game_fair} and
Figure~\ref{fig:cool_game_cyclic} respectively. These parameter vectors are
shown in Table~\ref{table:real_game_params}. We provide recordings of sample
episodes for each balancing scenario\footnote{Videos available at:
https://github.com/Danielhp95/GGJ-2020-cool-game }. As a game designer the most
important question to ask is: \textit{how do the different bots play?} We
briefly describe the game parameterized under $\bm{\theta}_{fair}$ and
$\bm{\theta}_{cyclic}$:

\subsubsection{Fair balancing}

Torch bot, with the most health (9), slowest movement (6) and lowest damage
(3), plays like a
tank\footnote{https://en.wikipedia.org/wiki/Tank\_(video\_games)}. Nail bot is
a ``glass cannon''; the fastest (2) and most damaging (7) character with the
lowest health (4). Its cooldown of 1 tick allows it to quickly react to
opponents close by, and to barrage other bots from a distance. Saw bot moves at
a medium speed 4 and has to carefully approach opponents, but once it reaches
them a victory is always guaranteed.

\subsubsection{Cyclic balancing}

Bot behaviours are similar to the previous case, with some differences. Saw bot
is slower (5), often using the stand still action (S) to time movement to avoid
damage. It exploits Torch bot's shorter range (3) and longer cooldown (5) to
wait for an opening from a distance Nail bot, as fast as before but even more
damaging (9) is able to position itself for a single nail that kills the slower
Saw bot. Because Nail bot now has only 3 health, it dies to a single touch by
Torch bot's flame, making it significantly weaker against it.

\subsection{Discussion}

\begin{figure}[htbp]
    \centering
    \includegraphics[width=0.8\linewidth,keepaspectratio]{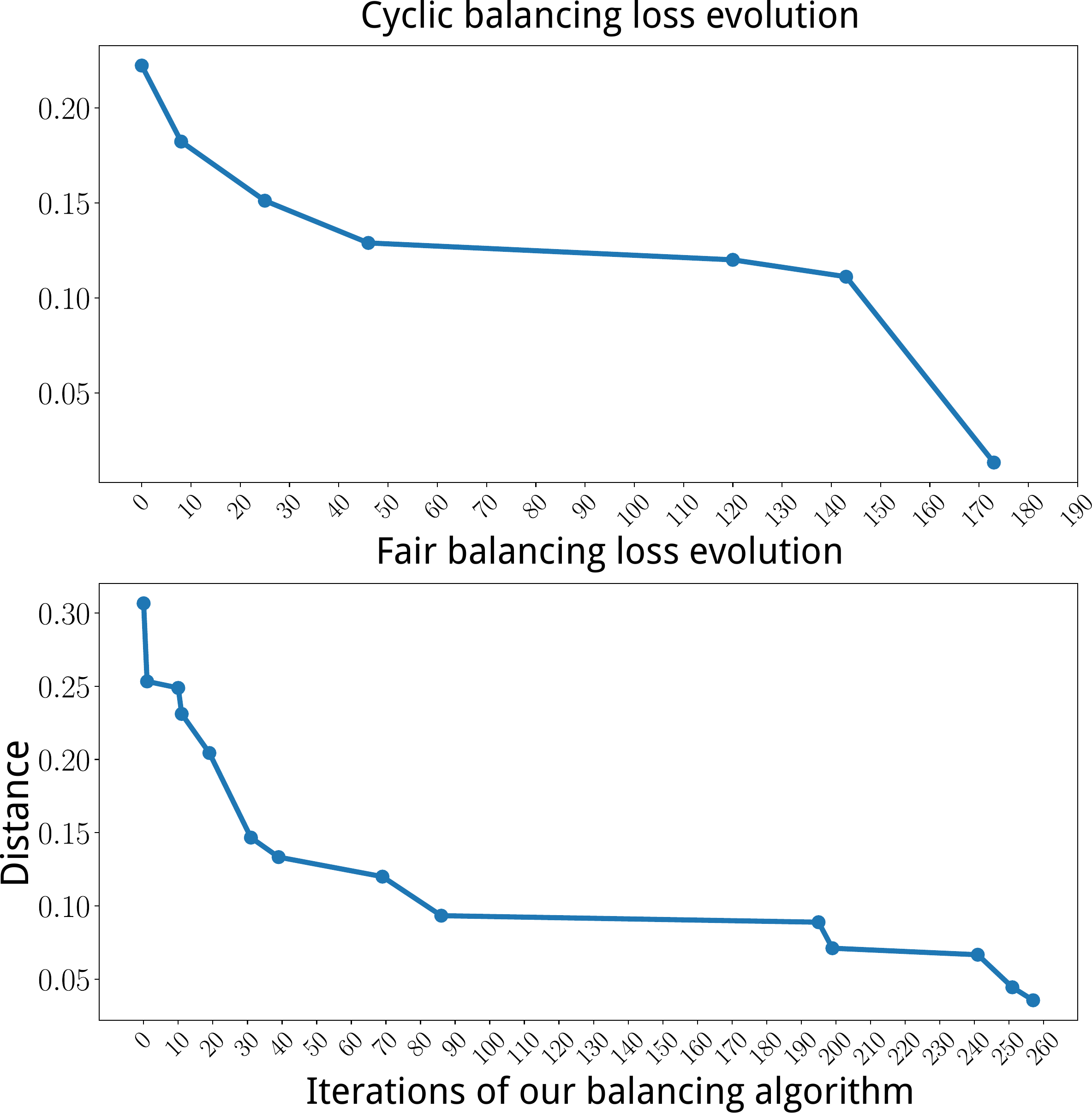}
    \caption{Evolution of distance to target balance graph for Cyclic balancing (top) and fair balancing (bottom). Top graph stops at iteration 173, as it was the last iteration to improve upon the previous best parameter.}
    \label{fig:cool_game_balancing_progression}
\end{figure}

Figure~\ref{fig:cool_game_balancing_progression} shows how, as our algorithm
iterated, both $\bm{\theta}_{fair}$ and $\bm{\theta}_{cyclic}$ generated game
balacings which grew closer to the desired target balances. At the end of the
260 iterations, The balancing emerging from $\bm{\theta}_{cyclic}$ features an
aggregated error of 9\% win-rate over the target graph which we deem as
acceptable. Unfortunately, the error associated with $\bm{\theta}_{fair}$ is
large (16\%) as the win-rate between Saw bot and Nail bot favoured Saw bot
heavily, which we deem as unsatisfactory. However, given the downwards trend of
Figure~\ref{fig:cool_game_balancing_progression}, we have reason to believe
that better paramater vectors could be found, provided greater computational
time.

For both experiments, each algorithmic iteration was completed, on average,
every 20-25 minutes, and in total both experiments took approximately 96 hours
each, where most of the computational time was spent by MCTS's internal
simulations. This is evidence that our algorithm is computationally expensive.
Although a linear speedup could be gained simply by increasing the number of
CPUs, further improvements aimed reducing the computational load of the algorithm
are needed to allow for the balancing of real-world games.

In Figure~\ref{fig:cool_game_balancing_progression} between iterations 0 and 80
there are 8 iterations, more or less evenly spaced, which improve upon the best
parameters found so far. Assuming each iteration takes 20 minutes, every 10
iterations, or roughly 3 hours and 20 minutes, our algorithm found game
parameters that moved the balancing closer to the designers' target balancing.
This is a clear example of how our algorithm automates the balancing process.
On the other hand, we also see in
Figure~\ref{fig:cool_game_balancing_progression} a gap between iterations 80
and 200 where our algorithm did not find a parameter vector which improved upon
the current best solution. In wall-clock time, this gap took 40h. This is
clearly an issue, specially for more computationally intensive games. From the
user's perspective, our method does not return any information during those 40h
because no new best parameter $\bm{\theta}$ was found. One is left to wonder if
there are any metrics not directly relevant to the optimization process, which
could be extracted from our algoritm's computation that may be of use to the
game developers. This remains an open question.

Certain parameterizations might defy the original intent of the designer. In
the field of AI, this is known as value misalignment. We name a few. In the
fair balancing case, all bots can die from either 1 or 2 hits, which makes for
short-lived matches. Furthermore, Saw bot's ability lasts for 6 ticks, whilst
having a low cooldown of just 3 ticks. This makes its ability an almost
permanent effect rather than a special action. As a parallel study, the
playstyle displayed by MCTS tends to be very offensive at the beginning, and
very defensive later on. More human-like methods for generating
gameplaying agents would greatly benefit the result of our algorithm (line 5 of
Algorithm~\ref{algorithm:automated_balancing}), although we emphasize that this
problem is orthogonal to our algorithm.

\begin{table}[h]
    \centering
    \caption{Optimized parameters of each bot type for fair and cyclic target graphs.}
    \label{table:real_game_params}
    \begin{tabular}{@{}llcccc@{}}
        \toprule
        Bot Type & Parameter & Min & Max & Fair & Cyclic\\
        \midrule
        Torch& Health& 1 & 10 & 9 & 7 \\
            &Cooldown& 1 & 6 & 3 & 5 \\
            &Damage& 1 & 10 & 3 & 3 \\
            &Ticks between move& 1 & 6 & 6 & 4 \\
            &Torch duration& 1 & 6 & 3 & 2 \\
            &Torch range& 1 & 4 & 4 & 3 \\
       \midrule
         Nail& Health& 1 & 10 & 4 & 3 \\
             &Cooldown& 1 & 6 & 1 & 1 \\
             &Damage& 1 & 10 & 7 & 9 \\
             &Ticks between move& 1 & 6 & 2 & 2 \\
        \midrule
         Saw& Health& 1 & 10 & 6 & 4 \\
             &Cooldown& 1 & 6 & 3 & 3 \\
             &Damage& 1 & 10 & 2 & 6 \\
             &Damage change& 1 & 10 & 7 & 6 \\
             &Ability duration& 1 & 6 & 6 & 3 \\
             &Ticks between move& 1 & 6 & 4 & 5 \\
        \bottomrule
    \end{tabular}
\end{table}

\begin{table}
    \centering
    \caption{Win rates for Saw Vs. Torch ($w_{ST}$), Saw Vs. Nail ($w_{SN}$) and Torch Vs. Nail ($w_{TN}$) after balancing and their corresponding errors.}
    \label{table:real_game_params}
    \begin{tabular}{@{} l rrr rrr @{}}
        \toprule
        & \multicolumn{3}{c}{\textbf{Fair balance: $\bm{\theta}_{fair}$}} & \multicolumn{3}{c}{\textbf{Cyclic balance: $\bm{\theta}_{cyclic}$}} \\
        \cmidrule(lr){2-4} \cmidrule(l){5-7}
                Match & Target & Found & Error           & Target & Found & Error\\
        \midrule
        $w_{ST}$ &   50\%  &  50\% & 0\%            &  70\%  &  68\% & 2\% \\ 
        $w_{SN}$ &   50\%  &  64\% & 14\%           &  30\%  &  36\% & 6\% \\ 
        $w_{TN}$ &   50\%  &  52\% & 2\%            &  70\%  &  69\% & 1\% \\ 
        \bottomrule
    \end{tabular}
\end{table}

\vspace{-1em}

\section{Related work}

Quantitative methods for understanding games have been proposed in many forms.
\cite{beyer2016integrated} presents a similar balancing overview to ours,
introducing a generic iterative an automated balancing process. of sampling
game parameters, using AI players (or real humans) and testing if a desirable
balancing has been achieved. Our main differentiating contribution are
balancing graphs as a designer friendly balancing description. Several
strategies for the assessment of games without real player data are described
in~\cite{Nelson2011GameMW}. Our research most closely resembles the strategy
defined as ``Hypothetical player-testing", in which we are ``trying to
characterize only how a game operates with a particular player
model"~\cite{Nelson2011GameMW}. The forms that this type of hypothetical
player-testing analysis can take are discussed in depth
in~\cite{jaffe2013understanding}, and specifically our work is concerned with
the subcategory of quantitative analysis defined as Automated Analysis, helping
designers evaluate (and often modify) their games without human
playtests~\cite{jaffe2013understanding}.

Machine Learning offers tools for automatic meta-game analysis. Harnessing
existing supervised learning algorithms, \cite{Argue2014SupervisedLA} used
random forests and different neural network architectures to assess meta-game
balance by predicting the outcomes of individual matches using hand-crafted
features that describe the strategies being used. The authors make an
assessment of the overall balance of the meta-game by measuring ``the
prevalence of parallel strategies"~\cite{Argue2014SupervisedLA}, assuming
balance to mean equal prevalence. While this is informative, it is predicated
on a definition of balance that may not align with the goals of other game
designers working on other projects, which may have definitions for balance
that extend beyond prevalence. Additionally, such techniques are limited to
assessing the current balance of a game context rather than providing a
solution for balance issues that are discovered.

MCTS has been used for this type of simulation based analysis in the past,
in~\cite{Zook2015MonteCarloTS} MCTS agents are used to model players at various
skill levels in Scrabble and Cardonomicon to extract metrics that describe game
balance. However, this type of analysis is concerned with the discovery of
issues and takes no steps towards providing a solution to a balance problem
once discovered.

The work by Liu and Marschner~\cite{Liu2017BalancingZG} uses the Sinkhorn-Knopp
algorithm to balance a mathematical model, according to game theoretical
constructs, representing a simplified version of the popular game Pokemon. In
Pokemon, each pokemon type\footnote{An overview of Pokemon types:
https://bulbapedia.bulbagarden.net/wiki/Type} has advantages and disadvantages
against various other types. The authors tune these type features to make them
all equally viable pokemon types. This is akin to our fair balancing experiment
in Section~\ref{section:experiments_real_game}. This approach concerns itself
with mathematical comparisons between strategies based on an existing table of
matchup statistics, which may not exist for most games, especially those still
in development.

Leigh et al.~\cite{leigh2008using} used co-evolution to evolve optimal
strategies for CaST, a capture the flag game. Populations of agents were
evolved in an environment with a set of game parameters. The distribution of
the resulting agents across simplex heat maps of different strategies was used
to assess whether or not the game was balanced with those game parameters by
considering balance to be a situation where any core strategy should beat one
of the other core strategies and lose against another, similar to our
definition of \textit{cyclic} balancing. They manually modified play parameters
and iterated to find a configuration with a desirable heatmap.
Our approach builds upon this work by automating the manual parameter
adjustment stage, it also broadens the definition of balance by allowing the
designer to specify exactly what meta-game state they consider balanced.

% A study on the Hearthstone meta-game has a similar approach to meta-game
% balance, ~\cite{de2019evolving} uses an evolutionary algorithm to search for a
% combination of changes to card attributes that create a hearthstone meta-game
% where every deck has a 50\% win-rate. Our work can be considered to be a
% generalization of this approach, which is not limited to Hearthstone and does
% not prescribe the precise conditions under which a game can be considered
% balanced.

%More general methods to evaluate balance and inform designers exist in the form
%of the restricted play work. ~\cite{AIIDE125470} uses agents with specific
%game-play restrictions, such as limitations on the specific moves that can be
%made by an agent, to provide insight into specific balance questions such as
%``How important is playing unpredictably?". This work highlights the need for
%designer input in the analysis process, as many definitions of balance are
%limited to assessments of similarity between strategies. These methods are well
%suited to use in conjunction with ours and could allow a designer to specify
%metrics beyond win-rate in the target balance graph, such as the frequency of
%specific moves.

\section{Conclusion}

In this paper we present an algorithm to autobalance a game as requested by a
designer. We do this by combining concepts from AI for gameplaying,
optimization, game and graph theory. We also develop the mathematical
foundation for this tool, demonstrating its empirical convergence in a simple
toy domain and showcasing its potential in a richer game environment. To our
knowledge, our work is one of the first steps in the field of game balancing
towards robust tools for automated balancing in multiagent games. The issues of
computational time, non human-like AI behaviour and the complexity of
generating gameplaying agents remain as obstacles in the path towards accessible
adoption of our algorithm by designers.

% One is to explore to what extent game mechanics can influence the meta-game of
% rational agents in more complex games. Other interesting aspect would be to use
% RL with Self-Play training (or other agent generating techniques) instead of
% MCTS for when MCTS is not good enough or when a model of the game is not
% available.

Our contributions could be transformed into the ``backend'' of an actual tool.
To make it amenable to be used by non-technical individuals, a user-friendly
``frontend'' should be developed, exposing an interface to (1) parameterize a
game and (2) make it easy to specify a level of abstraction and its
corresponding balance graph.

\bibliographystyle{IEEEtran}
\bibliography{main}
\end{document}